\pdfoutput=1
\documentclass[11pt]{article}

\usepackage{amsmath,amssymb}
\usepackage{booktabs}
\usepackage{cite}
\usepackage[margin=1in]{geometry}
\usepackage{graphicx}
\usepackage{float}
\usepackage{microtype}
\usepackage{multirow}
\usepackage{url}
\usepackage{listings}
\usepackage{xcolor}

\lstdefinestyle{prompt}{
  basicstyle=\ttfamily\scriptsize,
  breaklines=true,
  breakatwhitespace=false,
  columns=fullflexible,
  keepspaces=true,
  showstringspaces=false,
  frame=single,
  framerule=0.3pt,
  xleftmargin=3pt,
  xrightmargin=3pt,
  aboveskip=4pt,
  belowskip=5pt
}

\newcommand{\method}{NSF-LLM}
\newcommand{\cov}{\mathrm{Coverage}_{90}}

\title{Auditable Context-Aware HFMD Forecasting with Structured LLM Agents}
\author{
Joongwon Chae\textsuperscript{1} \quad
Runming Wang\textsuperscript{1} \quad
Chen Xiong\textsuperscript{1} \quad
Gong Yunhan\textsuperscript{1} \\
Lian Zhang\textsuperscript{4} \quad
Ji Jiansong\textsuperscript{2} \quad
Dongmei Yu\textsuperscript{2} \quad
Peiwu Qin\textsuperscript{2,3,$\dagger$} \\[0.7em]
\small \textsuperscript{1}Institute of Biomedicine and Health Engineering, \\
\small Tsinghua Shenzhen International Graduate School (SIGS),
Tsinghua University, Shenzhen, China \\
\small \textsuperscript{2}The Fifth Affiliated Hospital of Wenzhou Medical University,
Lishui 323000, China \\
\small \textsuperscript{3}Hengqin Laboratory, Zhuhai, China \\
\small \textsuperscript{4}The First Hospital of Hebei Medical University,
Shijiazhuang, China \\[0.5em]
\small \texttt{cai-zy24@mails.tsinghua.edu.cn} \\
\small \textsuperscript{$\dagger$}Corresponding author:
\texttt{pwqin@sz.tsinghua.edu.cn}
}
\date{}

\begin{document}
\maketitle

\begin{abstract}
Effective HFMD surveillance requires forecasts capturing both time-series patterns and contextual drivers such as school calendars, weather, and policy or surveillance reports. In clinical settings, forecasts must be trusted and actionable; thus, beyond point accuracy, decision-makers require concise, auditable explanations of why risk is expected to rise or fall. Classical models (e.g., ARIMA and Prophet) and foundation models (e.g., Chronos, Moirai, and TimesFM) treat external covariates as numerical inputs, lacking semantic reasoning to reflect epidemiological mechanisms or resolve conflicting signals. We propose a two-agent neuro-symbolic framework that decouples contextual interpretation from probabilistic forecasting. An LLM-based Event Interpreter ingests heterogeneous signals---school schedules, weather summaries, government reports, and clinical guidelines---and outputs a scalar transmission-impact signal. A Forecast Generator combines this signal with historical case counts to produce point forecasts that are mapped to probabilistic predictions through Poisson/negative-binomial moment matching. We focus on one-week-ahead rolling forecasts, aligning with weekly hospital-capacity planning and the rapid, context-driven inflections typical of HFMD. We evaluate on two datasets: Hong Kong surveillance (90 target weeks in 2023--2024) and Lishui hospital visits (33 target weeks in 2024). Against traditional and foundation-model baselines, our approach achieves competitive point accuracy while providing robust 90\% intervals (coverage approximately 0.85--1.00) and concise rationales. This demonstrates that integrating domain knowledge through LLM-based agents can match strong numerical forecasters while yielding interpretable, context-aware forecasts aligned with public-health decision-making. The project code will be publicly available upon acceptance.
\end{abstract}


\section{Introduction}
Hand, foot, and mouth disease (HFMD) is an acute pediatric enteroviral infection and a persistent public-health concern across the Asia-Pacific region \cite{ventarola2015update,xing2014hfmd,koh2016epidemiology}. Although most cases are mild, severe neurological complications motivate continuous surveillance \cite{ventarola2015update,xing2014hfmd}. HFMD transmission is shaped by region-specific seasonality, meteorological conditions, and school-based contact patterns \cite{koh2016epidemiology,wang2014beijing,zhu2023status,liu2013clusters}. With limited antiviral options, reliable short-horizon forecasting is therefore central to public-health planning \cite{lutz2019forecasting,zhao2023hybrid}.

In epidemic operations, a forecast is not merely a number. Clinicians and public-health staff need to understand which drivers, such as school attendance, weather anomalies, and surveillance reports, are responsible for anticipated changes, because decisions on staffing, triage readiness, and diagnostic suspicion are made under accountability constraints. This requirement is especially important for HFMD, which exhibits multi-peak seasonal patterns and abrupt contact-regime changes around school openings and holidays \cite{wang2014beijing,zhu2023status}. Effective forecasting therefore requires both temporal modeling and interpretation of heterogeneous contextual signals.

Existing forecasting approaches capture important parts of this problem but do not fully address it. Classical statistical models and machine-learning forecasters can model seasonal recurrence and nonlinear temporal patterns, while recent time-series foundation models provide strong pretrained numerical priors. However, these models typically treat external evidence as manually encoded covariates or raw numerical inputs, and they provide limited actionable rationales linking forecasts to concrete, domain-relevant drivers. This creates a bottleneck for clinical deployment, where users must assess not only the predicted count but also why risk is expected to rise or fall.

Large language models (LLMs) offer a natural interface for interpreting heterogeneous textual, categorical, and numerical context. At the same time, unrestricted LLM-based forecasting is undesirable in clinical settings because of hallucination risk, prompt sensitivity, and difficulty auditing free-form reasoning \cite{mccoy2025clinical,ravichander2025halogen,zhang2025hallucination}. We therefore use LLMs not as unconstrained numerical forecasters, but as bounded epidemiological context interpreters. The model is asked to convert disease-relevant evidence into a structured transmission-impact signal, confidence value, and concise rationale, which are then passed to a separate numerical forecasting and deterministic calibration layer.

We propose a two-agent neuro-symbolic framework in which an Event Interpreter processes external signals and maps them to a quantitative impact index $I_t$, while a Forecast Generator combines $I_t$ with historical case data to produce predictive distributions. This design preserves the strengths of numerical forecasting while adding an explicit, auditable reasoning layer that incorporates contemporaneous information at inference time. Our contributions are:
\begin{itemize}
\item A multi-agent neuro-symbolic architecture for HFMD forecasting that integrates contextual reasoning with probabilistic prediction. By decoupling semantic interpretation from numerical prediction, each component specializes while an inference-time adaptation mechanism incorporates contemporaneous context without retraining.
\item Structured interpretability through bounded impact scores, confidence values, and concise rationales that clarify how school calendars, weather patterns, and surveillance reports shape predicted outcomes.
\item Empirical validation on two real-world HFMD datasets with distinct temporal profiles, demonstrating competitive accuracy and calibrated uncertainty relative to classical, deep, and foundation-model baselines.
\end{itemize}

\section{Related Work}

\subsection{Epidemic and Time-Series Forecasting}
HFMD forecasting has been studied as a challenging epidemic time-series problem because transmission is shaped by seasonality, child contact patterns, and meteorological conditions. Prior epidemiological studies report that HFMD often exhibits a primary peak from late spring to early summer and a smaller secondary peak in early autumn, with substantial regional variation in seasonal intensity \cite{wang2014beijing,koh2016epidemiology,zhu2023status}. Temperature, humidity, and precipitation have nonlinear and lagged associations with HFMD activity, while school opening, holidays, and vacation periods can abruptly change contact opportunities among young children \cite{guo2016meteorological,hii2011weather,liu2013clusters,tian2018spatiotemporal}. These characteristics make HFMD forecasting more than a problem of extrapolating recent counts: short-horizon forecasts must also account for contemporaneous contextual drivers.

Classical epidemic and statistical forecasting methods, including SEIR-type models, ARIMA, Prophet, generalized additive models, and hybrid approaches, have been widely used for infectious-disease prediction \cite{kermack1927theory,hamilton2020timeseries,taylor2018prophet,xie2021prophet,liu2016arima}. These methods can capture seasonal recurrence and temporal dependence, and exogenous covariates can be incorporated when they are available in structured form. However, many operational signals in public-health forecasting are textual, categorical, delayed, or internally conflicting. For example, school reopening may increase contact, while extreme weather may temporarily suppress mobility. Standard numerical models can include such information as engineered covariates, but they do not directly expose the semantic reasoning used to reconcile these signals. This limits their usefulness when clinicians and public-health staff need an auditable explanation of why risk is expected to rise or fall.

\subsection{Machine Learning and Time-Series Foundation Models}
Machine-learning models have improved nonlinear time-series prediction through tree ensembles, recurrent neural networks, and Transformer-based architectures \cite{chen2016xgboost,ke2017lightgbm,prokhorenkova2018catboost,torres2021survey,nie2022patchtst,wu2022timesnet}. These models are effective when sufficient historical data are available and when the relevant predictive factors can be expressed as numerical features. More recently, time-series foundation models such as TimesFM, Chronos, and Moirai have shown that large pretrained models can provide strong numerical priors across diverse forecasting domains \cite{das2024timesfm,ansari2024chronos,woo2024unified}. Such models are valuable baselines for epidemic forecasting because they can capture broad temporal structure without training a disease-specific model from scratch.

Despite these strengths, time-series foundation models primarily operate on numerical histories or structured covariates. Their forecasts may track seasonal recurrence and recent momentum well, but they typically provide limited actionable rationales linking a predicted change to concrete public-health drivers. This distinction is important in clinical and surveillance settings. A model that accurately follows a stable seasonal curve may still be difficult to trust near abrupt transitions if it cannot explain how school calendars, weather anomalies, or regional bulletins influenced the forecast. Our work is therefore complementary to time-series foundation models: rather than replacing numerical forecasting, we introduce a structured contextual interpretation pathway that converts heterogeneous evidence into a bounded transmission-impact signal used by a probabilistic forecasting layer.

\subsection{LLMs for Contextual Evidence Integration}
Large language models provide a natural interface for integrating heterogeneous contextual evidence, including text reports, categorical event descriptions, and numerical summaries. Recent studies suggest that the role of LLMs in forecasting need not be limited to direct numerical prediction. TimeCAP uses LLM agents as contextualizers of time-series data, showing that generated contextual summaries can support downstream event prediction rather than replacing temporal models outright \cite{lee2025timecap}. Similarly, recent work on infectious-disease forecasting frames LLMs as tools for incorporating non-numerical information, such as public reports, policies, and surveillance evidence, into real-time forecasting workflows \cite{du2025pandemicllm}. These studies motivate our use of LLMs as bounded epidemiological context interpreters rather than unrestricted forecasters.

At the same time, simply providing more context to an LLM does not necessarily improve reasoning quality. Long-context analyses show that models may fail to reliably use relevant evidence when it is embedded in lengthy inputs, especially when the key information appears in the middle of the context \cite{liu2024lost}. Retrieval-augmented models can also be harmed by irrelevant or noisy retrieved passages, and recent RAG studies report that increasing the number of retrieved documents may initially help but later degrade generation quality because of hard negatives or redundant context \cite{yoran2024robust,jin2024longrag}. These findings motivate a selective evidence interface for epidemiological forecasting. Instead of prompting the model with entire guidance documents or raw surveillance reports, our framework retrieves a small number of disease-specific passages and converts external reports into compact evidence summaries. Given such targeted domain evidence, the LLM produces a bounded transmission-impact signal, confidence value, and concise rationale that can be audited before being passed to the numerical forecasting layer.

\subsection{Neuro-Symbolic and Agentic Forecasting}
Neuro-symbolic approaches combine the flexibility of statistical learning with the interpretability of symbolic structure \cite{yan2022neurosymbolic}. In time-series forecasting, this perspective is useful when numerical histories must be interpreted together with external rules, domain knowledge, or event descriptions. Epidemic forecasting is a natural setting for such a design because public-health decisions routinely combine observed incidence with school calendars, seasonal knowledge, weather conditions, surveillance reports, and clinical guidance.

Recent agentic LLM frameworks decompose complex tasks into structured reasoning and action steps, improving controllability and inspectability compared with unconstrained generation \cite{shinn2023reflexion,yao2023react,wang2023voyager,gui2025logicgame}. However, clinical applications also require caution: analyses of medical reasoning and hallucination emphasize the need for grounding, constrained outputs, and validation mechanisms \cite{mccoy2025clinical,ravichander2025halogen,zhang2025hallucination}. Our framework follows this principle by separating semantic interpretation, numerical forecasting, and probabilistic calibration. The Event Interpreter does not directly generate case counts; it maps evidence to a bounded impact score, confidence value, and concise risk notes. The Forecast Generator combines this structured signal with the historical incidence trajectory, and the final count distribution is constructed deterministically through Poisson or negative-binomial moment matching. This separation makes the contextual pathway auditable and allows each component to be ablated or replaced independently.

\section{Methods}
\subsection{System Overview}
\begin{figure*}[t]
\centering
\includegraphics[width=0.94\textwidth]{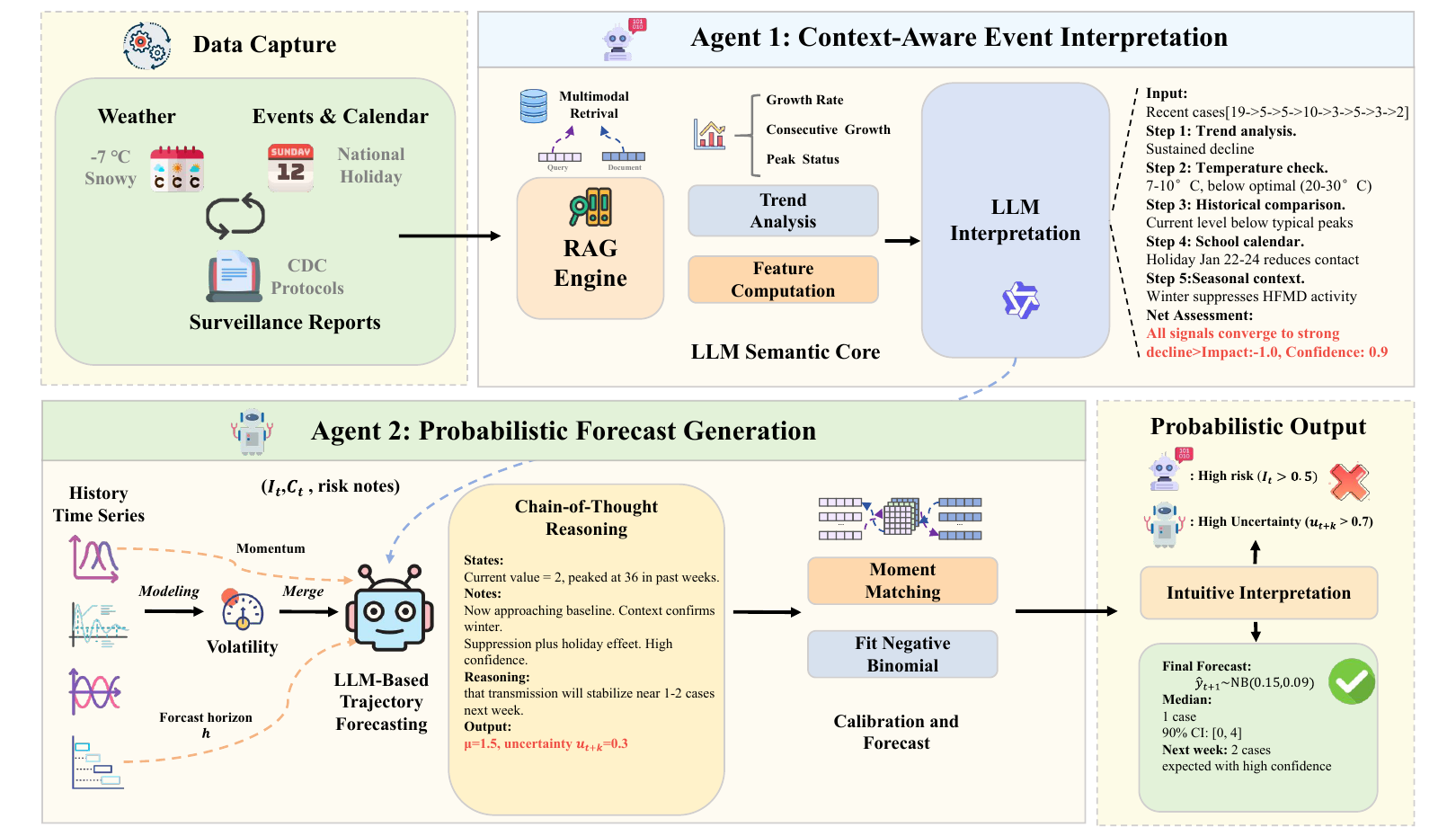}
\caption{Overall hierarchical neuro-symbolic architecture for HFMD forecasting. Agent 1 interprets heterogeneous contextual signals (school calendar, weather, regional HFMD trends, and guidelines) and emits a scalar transmission-impact signal, while Agent 2 combines this signal with historical case counts to produce probabilistic forecasts.}
\label{fig:overview}
\end{figure*}

We separate the epidemic time series and contextual evidence into distinct but coupled channels. The framework consists of two specialized agents. Agent 1, the Event Interpreter, processes weather summaries, school calendars, regional surveillance reports, and retrieved HFMD guidance. It returns a scalar transmission-impact signal $I_t\in[-1,1]$, a confidence score $C_t\in[0,1]$, an event summary, and short risk notes. Agent 2, the Forecast Generator, combines these outputs with the available incidence history and recent volatility to estimate the next-week mean count and uncertainty. The subsequent distributional calibration layer is deterministic and is not treated as a third agent. Fig.~\ref{fig:overview} shows the original two-agent architecture.

\subsection{Input Data Collection and Preprocessing}
Weather observations include mean temperature, relative humidity, and precipitation. Daily records are aggregated to epidemiological weeks to match the HFMD surveillance series. For each week, the seven-day average temperature, average humidity, and cumulative precipitation are serialized as structured inputs. School and social-event information is represented through categorical labels such as \texttt{in\_session}, \texttt{summer\_break}, \texttt{winter\_break}, and public holidays. Because HFMD predominantly affects young children, the interpreter is instructed to treat school status as a primary transmission driver rather than as a generic calendar covariate.

Regional surveillance information is extracted from official epidemiological bulletins. The original reports are unstructured; the pipeline therefore creates a lightweight summary containing recent monthly counts, month-over-month changes, and the relation between local hospital activity and broader regional activity. Each forecast origin receives only the latest bulletin that would have been available at that time. The resulting evidence pack preserves the distinction between local incidence, contemporaneous environmental conditions, and higher-level surveillance context. For example, a winter-holiday week is encoded separately from an in-session week, even when both have similarly low recent incidence. Likewise, a local rise can be interpreted differently when regional surveillance is simultaneously declining. These distinctions are retained in the structured prompt rather than collapsed into one aggregate feature before interpretation.

\subsection{Agent 1: Context-Aware Event Interpretation}
\subsubsection{Retrieval-Augmented Guidance}
Official HFMD diagnosis and management guidance is segmented into passages of at most 500 characters, embedded with the \texttt{all-MiniLM-L6-v2} encoder, and stored in a FAISS index. A dynamic retrieval query is composed from the current season, weather, and school status. For example, peak-season queries emphasize spring--summer activity, whereas winter queries emphasize low transmission; weather and school terms are appended when those signals are present. The top-$k$ passages ($k=2$, up to approximately 1200 characters) are inserted at the beginning of the interpreter prompt. This design anchors the assessment in disease-specific guidance without requiring the model to process a long undifferentiated document at every origin.

\subsubsection{Trend and Statistical Features}
Several lightweight statistics summarize the recent epidemic trajectory. Relative four-week growth is
\begin{equation}
 g_t=\frac{y_t-y_{t-4}}{\max(1.0,y_{t-4})}.
\end{equation}
The recent monotonic growth streak is
\begin{equation}
 w_{\mathrm{grow}}=\max\{k\mid y_{t-i+1}>y_{t-i},\ \forall i\in[1,k]\}.
\end{equation}
A peak indicator is set when the current observation exceeds the empirical 90th percentile of the historical series available at that origin:
\begin{equation}
 \mathrm{peak}_t=\mathbf{1}\!\left\{y_t\geq P_{90}(Y_{1:t})\right\}.
\end{equation}
The percentile is computed from the prefix available at the current origin, rather than from future observations. Together, growth, streak length, and peak status distinguish sustained movement from isolated fluctuations and provide Agent 1 with a compact description of the recent epidemic curve.

\subsubsection{Structured Output}
Agent 1 returns four fields: transmission impact $I_t\in[-1,1]$ (positive for promoting conditions and negative for suppressive conditions), confidence $C_t\in[0,1]$, a concise event summary, and short risk notes. Confidence is reduced when evidence is missing or when major signals point in different directions. The prompt requests a strict JSON object, allowing validation before the output is passed to Agent 2. The impact is defined for the expected effect beginning in the following week, consistent with the incubation and reporting delay used throughout the evaluation. Values near zero indicate that the available context does not justify a directional adjustment, whereas large absolute values are reserved for aligned evidence. The rationale is deliberately concise: it identifies the principal promoting and suppressive drivers without reproducing the full prompt or exposing unrestricted reasoning traces.

\subsubsection{Signal Reconciliation and Audit Interface}
The interpreter follows a disease-specific evidence hierarchy rather than treating all context fields as interchangeable covariates. School status is used as the primary behavioral signal because contact among kindergarten and primary-school children directly changes transmission opportunity. Temperature, humidity, and precipitation modulate whether the current season is favorable for sustained spread, while regional bulletins indicate whether the local trajectory is consistent with broader surveillance activity. The final impact score is therefore a semantic synthesis of these sources rather than a fixed linear combination. When school, weather, and surveillance signals agree, the magnitude of $I_t$ can increase together with confidence. When they disagree, the output preserves the competing directions in the risk notes and reduces $C_t$, allowing Agent 2 to retain stronger reliance on the observed trajectory.

This bounded interface serves two purposes. First, it makes the contextual adjustment inspectable: each weekly forecast can be traced to a compact impact score, confidence value, and evidence summary. Second, it separates interpretation from count generation. Agent 1 does not directly emit a case count, and Agent 2 does not independently reread the raw reports. Consequently, the same historical forecasting core can be evaluated with different interpreters, retrieval corpora, or context subsets while keeping the numerical task unchanged. The ablation study exploits this separation by removing climate, school-calendar, retrieval, or the complete Agent-1 pathway under otherwise identical forecast origins.

\subsection{Agent 2: Probabilistic Forecast Generation}
\subsubsection{Historical Volatility}
Recent volatility is estimated from the median absolute weekly relative change over the previous eight weeks:
\begin{equation}
 v_t=\operatorname{Clamp}\!\left(\operatorname{Median}_{j\in\mathcal{R}_t}
 \left|\frac{y_j-y_{j-1}}{\max(1.0,y_{j-1})}\right|,0.05,0.50\right).
\end{equation}
The median reduces sensitivity to isolated outliers, while clipping avoids unrealistically narrow or uninformative intervals.

\subsubsection{LLM-Based Trajectory Forecasting}
Agent 2 receives the full available history $Y_{1:t}$, the recent eight-week window, volatility $v_t$, Agent 1 outputs $(I_t,C_t)$ and risk notes, and the one-week horizon. It is instructed to summarize the long-term level and seasonal structure before interpreting the recent window, so that a short noisy swing does not override the multi-year pattern. The prediction is anchored in recent momentum and then adjusted according to the contextual signal and its confidence. When data momentum and $I_t$ conflict, the generator is instructed to preserve the observed trajectory unless the external evidence is coherent and high-confidence. The structured output contains a nonnegative mean forecast $\mu_{t+1}$ and an uncertainty score $u_{t+1}\in[0,1]$. Invalid or incomplete JSON outputs are reissued once under the same schema; no forecast origins were discarded in the reported experiments. The lag policy is applied consistently: recent values reflect the effects of earlier conditions, while the current impact score describes the expected net effect beginning at $t+1$. This distinction prevents a newly observed school event from being counted twice---once in the recent trajectory and again as an immediate unlagged adjustment.

\subsubsection{Distributional Calibration}
HFMD counts are discrete and often overdispersed. We therefore map $(\mu_{t+1},u_{t+1})$ to a count distribution by moment matching. The target variance is
\begin{equation}
 \sigma^2_{t+1}=\left[\mu_{t+1}v_t(1+u_{t+1})\right]^2.
\end{equation}
When $\sigma^2_{t+1}>\mu_{t+1}$, the negative-binomial parameters are
\begin{equation}
 n=\frac{\mu_{t+1}^2}{\sigma^2_{t+1}-\mu_{t+1}},\qquad
 p=\frac{n}{n+\mu_{t+1}}.
\end{equation}
Otherwise, a Poisson distribution is used. The 90\% interval is obtained from the 0.05 and 0.95 inverse CDF values. This construction prevents negative forecasts and allows asymmetric uncertainty at low counts. Increasing $u_{t+1}$ widens the target variance multiplicatively; for example, changing the uncertainty multiplier from 1 to 1.5 increases the variance contribution by a factor of $1.5^2=2.25$. The LLM therefore controls the degree of uncertainty, while the support, parameterization, and quantile computation remain deterministic. The final report can present the predictive mean, median, and 90\% interval without changing the role of $\mu_{t+1}$ in moment matching. This resolves the distinction between the mean generated by Agent 2 and the median optionally shown to users as a robust summary of the resulting count distribution.

\subsection{Data Sources, Ethics, and Execution Protocol}
The Lishui series contains 246 weekly outpatient-visit counts from January 2020 to September 2024, with rolling evaluation over 33 target weeks from February to September 2024. The Hong Kong archive contains 796 weekly HFMD-associated hospital-admission episodes from 2010 to 2025, with evaluation over 90 target weeks from January 2023 to September 2024. Within the evaluation windows, Lishui has mean weekly incidence 5.3, standard deviation 4.8, and maximum 19, while Hong Kong has mean 8.7, standard deviation 6.2, and maximum 31. Lishui context combines local meteorological records, the Chinese school calendar, and Zhejiang surveillance summaries; Hong Kong context combines Observatory weather records, Education Bureau calendars, and Centre for Health Protection reports.

Hong Kong surveillance data are publicly available. The Lishui dataset contains fully de-identified weekly aggregates provided under a data-sharing agreement; no individual-level identifiers were accessed or stored. The analysis uses only weekly counts and contextual summaries, and secondary analysis of anonymized aggregate data does not involve identifiable human-subject records under applicable rules.

At each forecast origin, the pipeline truncates all history and contextual records at that date, computes trend and volatility statistics, obtains $(I_t,C_t)$ from Agent 1, obtains $(\mu_{t+1},u_{t+1})$ from Agent 2, constructs the predictive distribution, and records the observed outcome. Evidence packs are assembled only from information available before the forecast date, including the preceding eight weeks of weather, active school or holiday status, and the most recent released surveillance summaries. The full prefix $Y_{1:t}$ is retained for seasonal comparison, while the recent eight-week window is used for momentum and volatility. This produces 33 Lishui and 90 Hong Kong forecasts without look-ahead leakage. Table~\ref{tab:param} summarizes fixed system parameters.

\begin{table}[t]
\caption{System parameters.}
\label{tab:param}
\centering
\footnotesize
\setlength{\tabcolsep}{4pt}
\begin{tabular}{lll}
\toprule
Parameter & Value & Description \\
\midrule
horizon & 1 week & Forecast window \\
recent window & 8 weeks & Trend/volatility lookback \\
volatility min & 0.05 & Minimum volatility \\
volatility max & 0.50 & Maximum volatility \\
\bottomrule
\end{tabular}
\end{table}

We use Qwen3-235B, GPT-5.1, and DeepSeek-V3.2 in thinking mode as LLM backbones. Each backbone uses one fixed decoding configuration throughout the study, with structured-output validation enabled for both agents. The maximum generation length is 2000 tokens, although the required JSON responses are substantially shorter. LLM-based configurations are repeated five times with identical inputs, and mean $\pm$ standard deviation is reported. Baselines are evaluated under the same rolling-origin history split through their native forecasting interfaces; external semantic context is processed by the proposed agent pathway. This comparison isolates the value of the structured context channel while preserving the standard operating mode of each baseline. ARIMA and Prophet are refit at each origin; LSTM and XGBoost use early stopping on the available training prefix; and the three foundation models use their official pretrained configurations. The same target dates and observations are used for every method. For probabilistic metrics, each implementation uses its corresponding predictive or uncertainty interface, and all interval construction is restricted to the training prefix at that origin.

\subsection{Evaluation Metrics}
Point accuracy is measured using mean absolute error (MAE) and root mean squared error (RMSE):
\begin{equation}
 \mathrm{MAE}=\frac{1}{N}\sum_{i=1}^{N}|y_i-\hat y_i|,\quad
 \mathrm{RMSE}=\sqrt{\frac{1}{N}\sum_{i=1}^{N}(y_i-\hat y_i)^2}.
\end{equation}
Probabilistic quality is assessed using the continuous ranked probability score (CRPS),
\begin{equation}
 \mathrm{CRPS}(F,y)=\int_{-\infty}^{\infty}
 \left(F(z)-\mathbf{1}\{z\geq y\}\right)^2\,dz,
\end{equation}
and empirical 90\% coverage,
\begin{equation}
 \cov=\frac{1}{N}\sum_{i=1}^{N}\mathbf{1}\{q_{0.05}^{(i)}\leq y_i\leq q_{0.95}^{(i)}\}.
\end{equation}
CRPS evaluates the complete predictive distribution and rewards both sharpness and proximity to the observation. Coverage assesses whether the nominal interval contains the realized count at the expected frequency. Lower MAE, RMSE, and CRPS are better, while coverage is interpreted relative to the nominal value of 0.90. MAE emphasizes typical absolute deviations, whereas RMSE gives greater weight to large misses around epidemic peaks. CRPS and coverage are therefore reported alongside the point metrics rather than replacing them, allowing the evaluation to distinguish center accuracy, tail behavior, and interval reliability.

\section{Experiments}

\subsection{Experimental Setup}
The experiments address four questions: whether structured contextual interpretation improves history-only forecasting (RQ1), how different LLM backbones behave under the same impact/confidence schema (RQ2), which contextual components contribute most to performance (RQ3), and whether the framework transfers across distinct surveillance regimes (RQ4). RQ1 is evaluated by removing Agent 1 while keeping the same trajectory generator and count calibration. RQ2 compares Qwen3-235B, GPT-5.1, and DeepSeek-V3.2-T under identical structured-output constraints. RQ3 removes climate evidence, school-calendar information, retrieved HFMD guidance, and the complete Event Interpreter separately. RQ4 compares the short, irregular Lishui hospital series with the longer Hong Kong surveillance archive.

Baselines include ARIMA, Prophet, LSTM, XGBoost, TimesFM, Chronos, and Moirai. Classical and learning baselines use rolling-origin fitting with hyperparameter selection restricted to the available prefix, while foundation models receive the case history available at each origin through their native forecasting interfaces. The proposed framework additionally receives the contemporaneous evidence pack.

Each forecast origin is processed independently using only information available at that date. Weather summaries, school status, and surveillance reports are aligned to the forecast origin and restricted by their public release dates, preventing look-ahead leakage. Metrics are aggregated over 33 Lishui and 90 Hong Kong origins. For LLM configurations, five repeated runs use identical historical and contextual inputs, so the reported standard deviation reflects generation variability rather than data-split variation.

The two datasets are intentionally used as complementary test beds: Lishui evaluates robustness under a short, low-count hospital series with irregular fluctuations, whereas Hong Kong evaluates performance under a longer surveillance archive with clearer recurrent seasonality. This design allows us to examine whether the same structured context interface remains useful when the temporal prior is weak, as in Lishui, or already strong, as in Hong Kong. All methods are evaluated on identical target dates and observations, ensuring that differences reflect forecasting behavior rather than evaluation-window variation.

\subsection{Prompt Templates}
\label{sec:prompts}

To improve reproducibility and make the LLM interface auditable, we use fixed structured prompts for both agents. The prompts are designed to enforce the HFMD lag policy, restrict the output space, and prevent the LLM from directly constructing the final count distribution. The following templates summarize the exact constraints used in all experiments.

In addition to the agent-specific instructions, both agents receive a short HFMD background note summarizing recurring epidemiological patterns: primary activity from late spring to early summer, a smaller autumn peak, reduced activity in winter, the importance of school-based contact among young children, favorable transmission under moderate temperature and humidity, and an approximate one-week reporting lag. These statements are provided as soft background knowledge rather than deterministic rules, and the agents are instructed to combine them with the observed history and evidence pack at each forecast origin.

\subsubsection{Agent 1: Event Interpreter Prompt}
\label{sec:prompt_agent1}

Agent 1 converts heterogeneous contextual evidence into a bounded transmission-impact signal. It is instructed to reason about the expected effect beginning at the next week, consistent with the incubation and reporting delay of HFMD. The prompt also prevents unrestricted narrative output by requiring a strict JSON object.

\begin{lstlisting}[style=prompt]
You are an infectious-disease analyst translating qualitative context into HFMD transmission signals.

IMPORTANT LAG POLICY:

* HFMD typically shows a 1-week delay between behavioral/environmental shifts and reported cases.
* The transmission_impact must describe the expected net effect starting next week (t+1), not primarily the current week.

INPUT:

* disease, date, horizon_weeks, impact_lag_weeks.
* recent_values ordered from old to new.
* trend statistics such as recent growth, growth streak, and peak status.
* external_data including school calendar, weather summary, surveillance reports, and retrieved HFMD guidance.

OUTPUT STRICT JSON ONLY:
{
"transmission_impact": float in [-1, 1],
"confidence": float in [0, 1],
"event_summary": "short natural-language summary",
"risk_notes": ["zero or more short bullet strings"],
"lag_rationale": "short note about lag/lead timing"
}

GUIDANCE:

1. Treat school status as the strongest behavioral driver, followed by temperature/humidity and surveillance reports.
2. Positive impact means conditions are expected to increase cases from t+1 onward; negative impact means suppressive pressure.
3. Reserve large absolute impact values for aligned evidence.
4. Reduce confidence when evidence is missing or signals conflict.
5. Do not output case counts.
6. Be concise and do not restate the full input.
   \end{lstlisting}

\subsubsection{Agent 2: Forecast Generator Prompt}
\label{sec:prompt_agent2}

Agent 2 receives the historical incidence trajectory and the structured output from Agent 1. It generates only a nonnegative mean forecast and an uncertainty score; the final probabilistic distribution is then constructed deterministically by the calibration layer.

\begin{lstlisting}[style=prompt]
You are assisting with weekly infectious-disease forecasting.

LAG POLICY:

* HFMD typically reacts to external events with an approximately 1-week delay.
* The week-1 forecast should be driven primarily by recent_values and recent trend, interpreted in the context of the full multi-year history.
* transmission_impact describes the expected net effect starting in week t + impact_lag_weeks.

READ INPUT IN THIS ORDER:

1. Full history:

* weekly dates and values over multiple years.
* summarize the overall level, seasonality, and how the current level compares with past years.

2. Recent 8-week window:

* recent_values and recent trend statistics.
* use this as a zoom-in on local momentum, but do not let short noisy swings override long-term seasonal structure.

3. Event Interpreter output:

* transmission_impact in [-1, 1].
* confidence in [0, 1].
* risk_notes explaining school, weather, surveillance, or policy drivers.
* historical_volatility.

4. Forecast configuration:

* horizon_weeks.
* impact_lag_weeks.

OUTPUT STRICT JSON ONLY:
{
"forecast_mean": [float >= 0],
"uncertainty_scale": float in [0, 1],
"rationale": "English explanation, no more than 3 sentences, mentioning the lag policy"
}

CONSTRAINTS:

1. Do not output quantiles or a probability distribution.
2. forecast_mean must have length equal to horizon_weeks.
3. All forecast_mean values must be nonnegative.
4. uncertainty_scale must be a single float in [0, 1].
5. Avoid implausible spikes that contradict the lag policy, long-term seasonality, or historical volatility.
6. Distinguish effects already visible in recent_values from expected effects of current conditions through transmission_impact.
   \end{lstlisting}

\subsection{Main Quantitative Results}
Table~\ref{tab:lishui} reports results on Lishui. TimesFM achieves the lowest MAE (3.972), while the Qwen3-based framework achieves the lowest RMSE (5.688) and the highest empirical coverage (1.000). Moirai obtains the lowest CRPS. These results indicate that the proposed framework is strongest at reducing large misses and maintaining reliable intervals, while remaining close to the best point-error baseline.

\begin{figure}[t]
\centering
\includegraphics[width=0.99\textwidth]{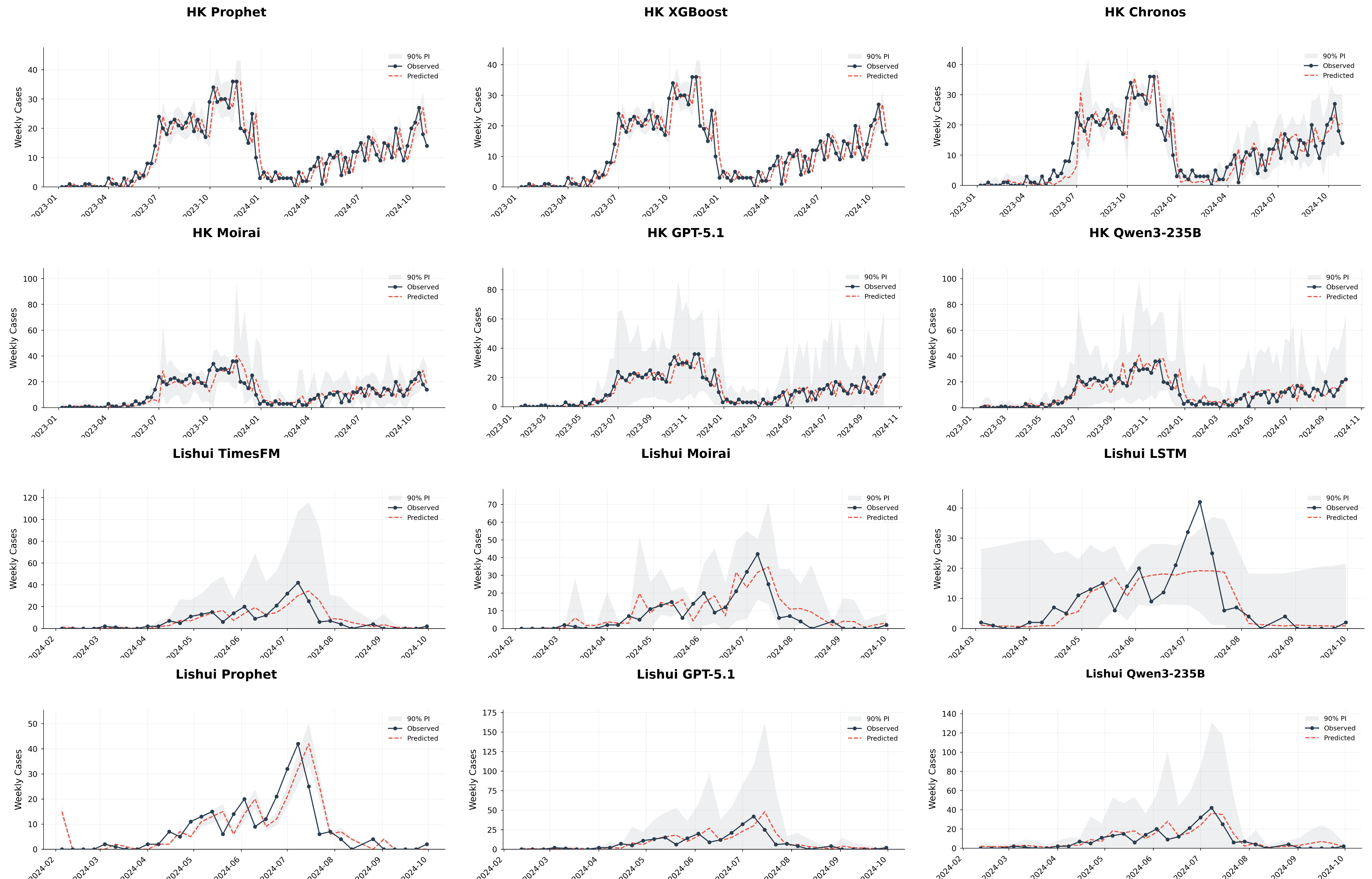}
\caption{Representative one-step-ahead trajectory panels for selected baselines and LLM-agent variants on Hong Kong and Lishui. Each panel shows observed counts, predicted means, and 90\% prediction intervals when available; full residual and density diagnostics are provided in the supplementary material.}
\label{fig:trajectories}
\end{figure}

Table~\ref{tab:hk} shows Hong Kong results. Under stable seasonality, Prophet and XGBoost achieve the lowest MAE (3.49), but their empirical coverage remains low. The GPT-5.1 variant matches this MAE, obtains the lowest RMSE (4.815), and reaches substantially higher coverage (0.879). Qwen3-235B obtains the highest coverage (0.912), followed by DeepSeek-V3.2-T (0.901). Overall, classical and foundation-model baselines remain strong numerical forecasters, while the LLM-agent variants occupy a complementary operating region with competitive point error and more reliable interval coverage. Figure~\ref{fig:trajectories} summarizes representative trajectories for selected baselines and LLM-agent variants.

\begin{table}[t]
\caption{One-step-ahead results on Lishui (33 weeks). Lower is better for MAE, RMSE, and CRPS.}
\label{tab:lishui}
\centering
\scriptsize
\setlength{\tabcolsep}{2.6pt}
\begin{tabular}{lcccc}
\toprule
Model & MAE & RMSE & CRPS & Coverage \\
\midrule
\method{} (Qwen3-235B) & 4.124$\pm$.31 & \textbf{5.688$\pm$.39} & 2.319$\pm$.11 & \textbf{1.000$\pm$.00} \\
\method{} (GPT-5.1) & 4.421$\pm$.37 & 7.058$\pm$.46 & 2.525$\pm$.13 & .848$\pm$.03 \\
\method{} (DeepSeek-V3.2-T) & 5.629$\pm$.53 & 9.732$\pm$.61 & 2.955$\pm$.16 & .939$\pm$.02 \\
TimesFM & \textbf{3.972} & 5.930 & 2.576 & 0.941 \\
Chronos & 4.403 & 6.630 & 2.105 & 0.412 \\
Moirai & 4.844 & 6.300 & \textbf{1.785} & 0.794 \\
LSTM & 4.463 & 6.862 & 2.621 & 0.933 \\
ARIMA & 4.853 & 7.374 & 3.935 & 0.265 \\
Prophet & 4.676 & 6.930 & 3.501 & 0.324 \\
XGBoost & 4.353 & 6.472 & 3.490 & 0.260 \\
\bottomrule
\end{tabular}
\end{table}

\begin{table}[t]
\caption{One-step-ahead results on Hong Kong (90 weeks).}
\label{tab:hk}
\centering
\scriptsize
\setlength{\tabcolsep}{2.6pt}
\begin{tabular}{lcccc}
\toprule
Model & MAE & RMSE & CRPS & Coverage \\
\midrule
Prophet & \textbf{3.49} & 4.83 & 2.31 & 0.400 \\
XGBoost & \textbf{3.49} & 4.83 & 2.53 & 0.337 \\
ARIMA & 3.51 & 4.83 & 2.54 & 0.326 \\
LSTM & 3.53 & 5.15 & 1.79 & 0.737 \\
TimesFM & 3.67 & 6.66 & 2.44 & 0.859 \\
Chronos & 3.79 & 5.23 & \textbf{1.71} & 0.568 \\
Moirai & 3.95 & 5.73 & 1.86 & 0.832 \\
\method{} (GPT-5.1) & \textbf{3.49$\pm$.24} & \textbf{4.815$\pm$.28} & 2.08$\pm$.09 & .879$\pm$.02 \\
\method{} (Qwen3-235B) & 3.95$\pm$.28 & 5.791$\pm$.35 & 2.33$\pm$.10 & \textbf{.912$\pm$.02} \\
\method{} (DeepSeek-V3.2-T) & 4.32$\pm$.34 & 5.87$\pm$.36 & 2.40$\pm$.12 & .901$\pm$.02 \\
\bottomrule
\end{tabular}
\end{table}

\subsection{Ablation Study}
Table~\ref{tab:ablation} isolates contextual components on Lishui. Removing Agent 1 degrades all metrics, confirming that the structured context pathway contributes beyond the shared forecasting core. The largest degradation occurs without school-calendar information, where MAE increases from 4.124 to 6.695 and coverage falls from 1.000 to 0.519, consistent with the role of child contact patterns in HFMD transmission. Climate removal mainly affects distributional quality and interval reliability, while removing retrieved guidance reduces both point and probabilistic performance. Because the no-Agent-1 variant retains the same trajectory generator and deterministic calibration layer, its degradation reflects the loss of the explicit context interface rather than a change in the numerical backend.

\begin{table}[H]
\caption{Ablation on Lishui (Qwen3-235B backbone).}
\label{tab:ablation}
\centering
\footnotesize
\setlength{\tabcolsep}{2.7pt}
\begin{tabular}{lcccc}
\toprule
Configuration & MAE & RMSE & CRPS & Coverage \\
\midrule
Full system & \textbf{4.124} & \textbf{5.688} & \textbf{2.319} & \textbf{1.000} \\
No Agent 1 & 4.621 & 6.984 & 2.660 & 0.879 \\
No climate & 4.424 & 7.049 & 3.692 & 0.788 \\
No RAG & 4.818 & 8.465 & 3.769 & 0.879 \\
No school calendar & 6.695 & 11.813 & 5.654 & 0.519 \\
\bottomrule
\end{tabular}
\end{table}

\subsection{Case Study: Seasonal Inflection}
We analyze 8 July 2024, when incidence had risen sharply, warm and humid weather remained favorable, and the school summer break introduced a suppressive factor. The observed count one week later was 25. As Table~\ref{tab:case} shows, Qwen3-235B balanced recent momentum against moderate school-break suppression, GPT-5.1 retained stronger short-term momentum under the lag rule, and DeepSeek-V3.2-T assigned the strongest suppressive impact. Their forecasts of 35, 48, and 46 all exceeded the observation, but the differences are traceable to the strength and timing assigned to the same school-break evidence rather than to changes in the deterministic calibration layer. This illustrates the practical value of separating event interpretation from distribution construction. This case highlights the role of the structured interface. The three backbones observe the same recent trajectory and external evidence, but express different assumptions through the bounded impact score and lag rationale. Because the calibration layer is fixed, the disagreement among forecasts can be attributed to contextual interpretation rather than to a hidden change in the probabilistic model. This makes the error analysis more transparent than a single black-box numerical forecast.

\begin{table}[H]
\caption{Interpretations and one-week-ahead forecasts for the 8 July 2024 inflection point.}
\label{tab:case}
\centering
\footnotesize
\setlength{\tabcolsep}{4pt}
\renewcommand{\arraystretch}{1.15}
\begin{tabular}{p{1.8cm}cp{4.3cm}p{5.0cm}c}
\toprule
Model & Impact & School-break signal & Weather/trend signal & Forecast \\
\midrule
Qwen3 & $-0.4$ &
Moderate suppression from the following week. &
Warm, humid weather supports transmission; upward pressure is partly mitigated. &
35 \\
GPT-5.1 & $-0.4$ &
Closure reduces transmission, but lag limits its immediate effect. &
Weather and recent momentum dominate the next-week trajectory. &
48 \\
DS-V3.2-T & $-0.6$ &
Strong suppressive factor beginning at $t+1$. &
Favorable weather partly offsets closure; growth is moderated. &
46 \\
\bottomrule
\end{tabular}
\end{table}

\section{Conclusion}
We proposed a hierarchical two-agent framework that augments HFMD forecasting with contextual reasoning from large language models. The system interprets weather, school calendars, retrieved guidance, and surveillance reports as a structured transmission-impact signal, combines it with historical incidence, and constructs Poisson or negative-binomial predictive distributions through moment matching. Across Lishui and Hong Kong, the framework achieves competitive point accuracy, strong empirical coverage, and clear gains from its contextual components. These results support LLMs as context interpreters within probabilistic biomedical forecasting systems. The impact score is directional rather than causal; future work will test broader diseases, regions, and horizons.
\bibliographystyle{unsrt}
\bibliography{references}

\end{document}